\pdfoutput=1

\documentclass[11pt]{article}

\usepackage[final]{acl}

\usepackage{times}
\usepackage{latexsym}

\usepackage[T1]{fontenc}

\usepackage[utf8]{inputenc}

\usepackage{algorithm}
\usepackage{algpseudocode}
\usepackage{amsmath,amsfonts,amssymb}
\usepackage{booktabs}
\usepackage{enumitem}
\usepackage{graphicx}
\usepackage{inconsolata}
\usepackage{microtype}
\usepackage{paralist}
\usepackage{xcolor}
\usepackage{xspace}

\newcommand{\fe}[1]{\texttt{#1}}
\newcommand{\framet}[1]{\texttt{#1}}
\newcommand{\gpt}{\texttt{GPT-4}\xspace}
\newcommand{\gptfe}{\texttt{GPT-4} | FE\xspace}
\newcommand{\gptframefe}{\texttt{GPT-4} | Frame+FE\xspace}
\newcommand{\tfive}{\texttt{T5}\xspace}
\newcommand{\tfivefe}{\texttt{T5} | FE\xspace}
\newcommand{\tfiveframefe}{\texttt{T5} | Frame+FE\xspace}

\newcommand{\draftonly}[1]{#1}
\renewcommand{\draftonly}[1]{}

\title{Annotating FrameNet via Structure-Conditioned Language Generation}

\author{Xinyue Cui \\
  University of Southern California \\
  \texttt{xinyuecu@usc.edu} \\\And
  Swabha Swayamdipta \\
  University of Southern California \\
  \texttt{swabhas@usc.edu} \\}

\begin{document}
\maketitle

\begin{abstract}
Despite the remarkable generative capabilities of language models in producing naturalistic language, their effectiveness on explicit manipulation and generation of linguistic structures remain understudied. 
In this paper, we investigate the task of generating new sentences preserving a given semantic structure, following the FrameNet formalism. 
We propose a framework to produce novel frame-semantically annotated sentences following an overgenerate-and-filter approach. 
Our results show that conditioning on rich, explicit semantic information tends to produce generations with high human acceptance, under both prompting and finetuning.
Our generated frame-semantic structured annotations are effective at training data augmentation for frame-semantic role labeling in low-resource settings; however, we do not see benefits under higher resource settings. 
Our study concludes that while generating high-quality, semantically rich data might be within reach, the downstream utility of such generations remains to be seen, highlighting the outstanding challenges with automating linguistic annotation tasks.\footnote{Our code is available at \url{https://github.com/X-F-Cui/FrameNet-Conditional-Generation}.
}
\end{abstract}

\begin{figure}[t!]
\centering
\includegraphics[width=0.5\textwidth]{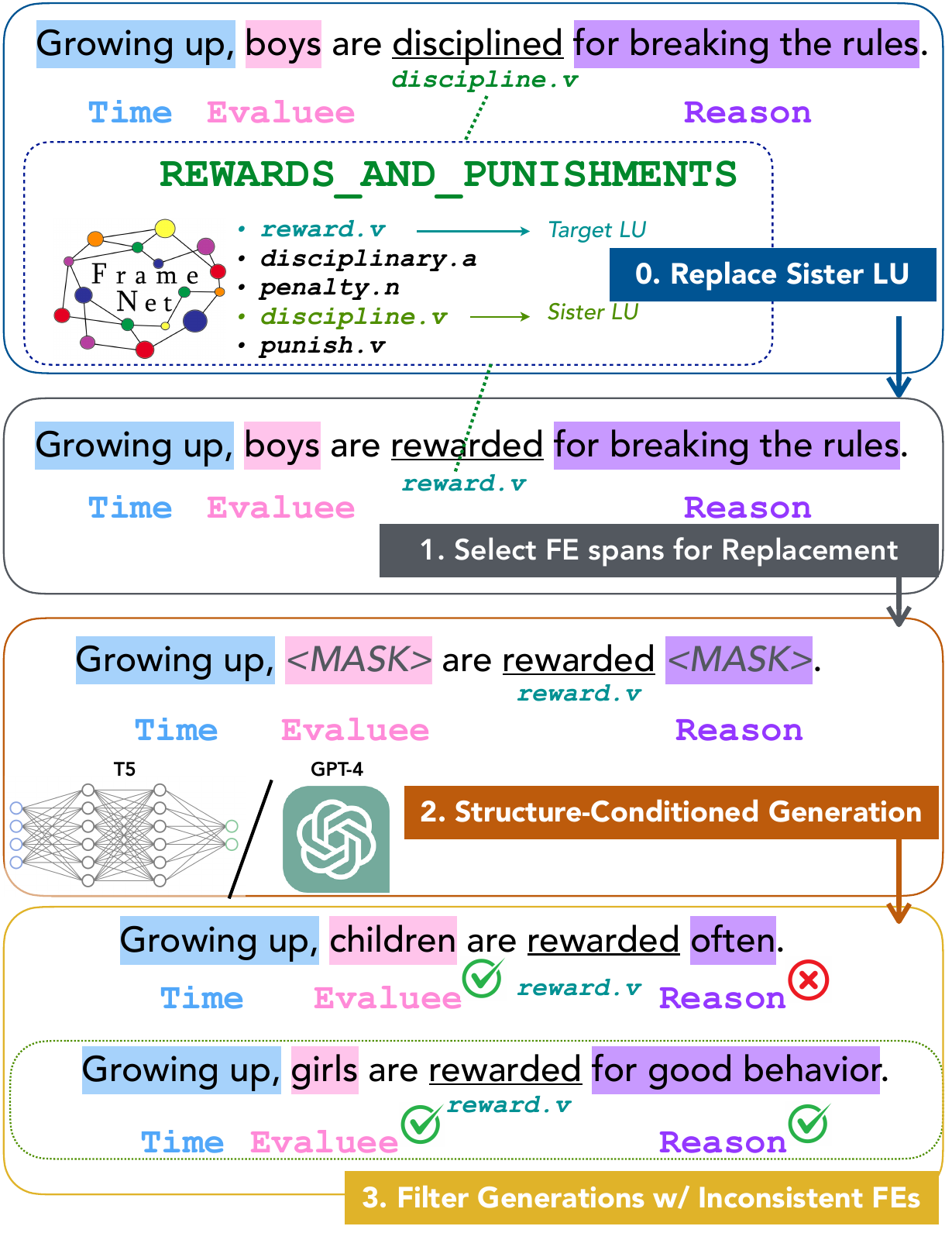}
\caption{Our framework to generate frame semantic annotated data. 
Following \citet{Pancholy2021SisterHD}, we replace a sister LU with the target LU in an annotated sentence (0;\S\ref{sec:sister-replace}). 
We select FEs appropriate for generating a new structure-annotated sentence (1;\S\ref{sec:method-masking}), and execute generation via fine-tuning T5 or prompting \gpt (2;\S\ref{sec:method-generation}). Finally, we filter out sentences that fail to preserve LU-FE relationships under FrameNet (3;\S\ref{sec:method-filtering}).
}
\label{figure:pipeline}
\end{figure}

\section{Introduction}

Large language models (LLMs) have demonstrated unprecedented capabilities in generating naturalistic language. 
These successes hint at LMs' implicit capabilities to \textit{``understand''} language; but are they capable of processing explicit symbolic structures in order to generate language consistent with the structures?
Not only would this help us understand the depth of LLMs' linguistic capabilities but would also serve to efficiently and cheaply expand existing  sources of linguistic structure annotation.
In this work, we investigate the abilities of LLMs to generate annotations for one such resource of linguistic structure: FrameNet, a lexical database grounded in the theory of frame semantics \citep{Fillmore:85,Ruppenhofer:16}.
We propose an approach for language generation conditioned on frame-semantic structure such that the generation (i) is consistent with the frame structure, (ii) is acceptable by humans and (iii) is useful for a downstream task, namely frame-semantic role labeling \cite{gildea-jurafsky-2000-automatic}. 

Our framework for generating frame-semantic annotations leverages both the FrameNet hierarchy and LLMs' generative capabilities to transfer annotations from existing sentences to new examples. 
Specifically, we introduce frame structure-conditioned language generation, focused on specific spans in the sentence such that the resulting sentence follows the given frame structure and is also acceptable to humans.
Overall, we follow an overgenerate-and-filter pipeline, to ensure semantic consistency of the resulting annotations.
Our framework is outlined in \autoref{figure:pipeline}.

Our intrinsic evaluation, via both human judgment and automated metrics, show that the generated sentences preserve the intended frame-semantic structure more faithfully compared to existing approaches \cite{Pancholy2021SisterHD}.
As an extrinsic evaluation, we use our generations to augment the training data for frame-semantic role labeling: identifying and classifying spans in the sentence corresponding to FrameNet frames.
Under a low-resource setting, our generation annotations tend to be effective for training data augmentation for frame-semantic role labeling.
However, these trends do not translate to a high-resource setting; these findings are consistent with observations from others who have reported challenges in leveraging LLMs for semantic parsing tasks, such as constituency parsing \cite{Bai2023ConstituencyPU}, dependency parsing \cite{Lin2023ChatGPTIA}, and abstract meaning representation parsing \cite{Ettinger2023YouAA}. 
Our findings prompt further investigation into the role of LLMs in semantic structured prediction.

\section{FrameNet and Extensions}
\label{sec:framenet-bg}

Frame semantics theory \cite{Gildea2000AutomaticLO} posits that understanding a word requires access to a \textbf{semantic frame}---a conceptual structure that represents situations, objects, or actions, providing context to the meaning of words or phrases. 
\textbf{Frame elements (FEs)} are the roles involved in a frame, describing a certain aspect of the frame. 
A \textbf{Lexical Unit (LU)} is a pairing of tokens (specifically a word lemma and its part of speech) and their evoked frames.
As illustrated in \autoref{figure:pipeline}, the token ``disciplined'' evokes the LU \textit{discipline.v}, which is associated with the frame \framet{REWARDS\_AND\_PUNISHMENT}, with FEs including \fe{Time}, \fe{Evaluee}, and \fe{Reason}.
Grounded in frame semantics theory, FrameNet \cite{Ruppenhofer2006FrameNetIE} is a lexical database, featuring sentences that are annotated by linguistic experts according to frame semantics. 
Within FrameNet, the majority of sentences are annotated with a focus on a specific LU within each sentence, which is referred to as lexicographic data; \autoref{figure:pipeline} shows such an instance. 
A subset of FrameNet's annotations consider all LUs within a sentence; these are called full-text data; \autoref{figure:pipeline} does not consider other LUs such as \textit{grow.v} or \textit{break.v}.

FrameNet has defined 1,224 frames, covering 13,640 lexical units.
The FrameNet hierarchy also links FEs using 10,725 relations.
However, of the 13,640 identified LUs, only 62\% have associated annotations.
Our approach seeks to automatically generate annotated examples for the remaining 38\% of the LUs, towards increasing coverage in FrameNet without laborious manual annotation.

\paragraph{Sister LU Replacement}
\label{sec:sister-replace}
\citet{Pancholy2021SisterHD} propose a solution to FrameNet's coverage problem using an intuitive approach: since LUs within the same frame tend to share similar annotation structures, they substitute one LU (the \textbf{target LU}) with another (a \textbf{sister LU}) to yield a new sentence. 
This replacement approach only considers LUs with the same POS tag to preserve the semantics of the original sentence; for instance, in \autoref{figure:pipeline}, we replace the sister LU \textit{discipline.v} with the target LU \textit{reward.v}. 
However, due to the nuanced semantic differences between the two LUs, the specific content of the FE spans in the original sentence may no longer be consistent with the target LU in the new sentence.
Indeed \citet{Pancholy2021SisterHD} report such semantic mismatches as their primary weakness. 

To overcome this very weakness, our work proposes leveraging LLMs to generate FE spans that better align with the target LU, as described subsequently.
For the rest of this work, we focus solely on verb LUs, where initial experiments showed that the inconsistency problem was the most severe. 
Details of FrameNet's LU distribution by POS tags, along with examples of non-verb LU replacements can be found in \autoref{app:fn-pos-stats}.

\section{Generating FrameNet Annotations via Frame-Semantic Conditioning} 
\label{sec:method}

We propose an approach to automate the expansion of FrameNet annotations by generating new annotations with language models.
Given sister LU-replaced annotations (\S\ref{sec:sister-replace}; \citealp{Pancholy2021SisterHD}), we select FE spans which are likely to be semantically inconsistent (\S\ref{sec:method-masking}), generate new sentences with replacement spans by conditioning on frame-semantic structure information (\S\ref{sec:method-generation}) and finally filter inconsistent generations (\S\ref{sec:method-filtering}).

\subsection{Selecting Candidate FEs for Generation} 
\label{sec:method-masking}

We identify the FEs which often result in semantic inconsistencies, in order to generate replacements of the spans corresponding to such FEs.
Our selection takes into account the FE type, its ancestry under FrameNet, and the span's syntactic phrase type.
Preliminary analyses, detailed in \autoref{app:candidate-fe}, help us narrow the criteria as below:  
\begin{compactenum}
    \item \textbf{FE Type Criterion}: The FE span to be generated  must belong to a core FE type, i.e., the essential FEs that are necessary to fully understand the meaning of a frame.
    \item \textbf{Ancestor Criterion}: The FE should not possess Agent or Self-mover ancestors.
    \item \textbf{Phrase Type Criterion}: The FE's phrase type should be a prepositional phrase.
\end{compactenum}

Qualitative analyses revealed that it suffices to meet criterion (1) while satisfying either (2) or (3). 
For instance, in \autoref{figure:pipeline}, under \framet{REWARDS\_AND \_PUNISHMENTS}, only the FEs \fe{Evaluee} and \fe{Reason} are core (and satisfy (2)) while \fe{Time} is not; thus we only select the last two FE spans for generation. 

\subsection{Generating Semantically Consistent Spans} 
\label{sec:method-generation}

We generate semantically consistent FE spans for selected candidate FEs via two approaches: finetuning a \texttt{T5-large}  model \cite{Raffel2019ExploringTL} and prompting \gpt\texttt{Turbo}, following \citet{Mishra2021CrossTaskGV}.
In each case, we condition the generation on different degrees of semantic information:\\
\textbf{No Conditioning} \quad We generate FE spans without conditioning on any semantic labels. 
\\
\textbf{FE-Conditioning} \quad The generation is conditioned on the type of FE span to be generated. 
\\
\textbf{Frame+FE-Conditioning} \quad The generation is conditioned on both the frame and the FE type.

The above process produces new sentences with generated FE spans designed to align better with the target LU, thereby preserving the original frame-semantic structure. 
However, despite the vastly improved generative capabilities of language models, they are still prone to making errors, thus not guaranteeing the semantic consistency we aim for.
Hence, we adopt an overgenerate-and-filter approach \cite{langkilde-knight-1998-generation-exploits,walker-etal-2001-spot}: generate multiple candidates and aggressively filter out those that are semantically inconsistent.
Details on fine-tuning \tfive and prompting \gpt are provided in \autoref{app:t5-tempplates}.

\subsection{Filtering Inconsistent Generations} 
\label{sec:method-filtering}
\begin{table*}[ht!]
\centering
\small
\begin{tabular}{lrrrrr}
\toprule
&    \multicolumn{3}{c}{\bf Before Filtering ($|D_{\text{test}}|$=1K)} & \multicolumn{2}{c}{\bf After Filtering (FE Fid. = 1.0)} \\
\cmidrule(lr){2-4}\cmidrule(lr){5-6}
&     \bf FE Fid.   & \bf ppl.  & \bf Human ($|D_{\text{test}}|$=200)  &    \bf ppl.($|D_{\text{test}}|$) & \bf Human ($|D_{\text{test}}|$)             \\ 
\midrule
Human (FN 1.7) & 0.979  & 78.1  & 1.000  &    97.0 (975)  & 1.000 (199)  \\ 
Pancholy et al. & 0.953  & 127.8  & 0.611    & 146.0 (947)   & 0.686 (189)  \\ 
\midrule[0.03em]
\texttt{T5} & 0.784 & 139.3 & 0.594   & 117.5 (789)& 0.713 (156)  \\
\texttt{T5} | FE & 0.862 & 127.6 & 0.711    & 112.7 (850)  & 0.777 (168)  \\
\texttt{T5} | Frame + FE & \textbf{0.882} & 136.8 & 0.644   & 124.4 (\textbf{873})  & 0.704 (\textbf{172})  \\ 
\midrule[0.03em]
\texttt{GPT-4} & 0.704 & 114.9 & 0.528    & 114.2 (724)  & 0.723 (132)  \\
\texttt{GPT-4} | FE & 0.841 & \textbf{106.3} & 0.700   & \textbf{103.4} (838)  & \textbf{0.826} (164)  \\
\texttt{GPT-4} | Frame + FE & 0.853 & 117.2 & \textbf{0.733}  & 111.8 (845)  & 0.821 (165) \\
\bottomrule
\end{tabular}
\caption{Perplexity, FE fidelity and human acceptability of \texttt{T5} and \texttt{GPT-4} generations conditioned on different degrees of semantic information. 
Number of instances after filtering are in parantheses.
Best results are in boldface.
}
\label{tab:intrinsic}
\end{table*}

We design a filter to ensure that the generated sentences preserve the same semantics as the expert annotations from the original sentence.
This requires the new FE spans to maintain the same FE type as the original. 
We propose a new metric \textbf{FE fidelity}, which checks how often the generated spans have the same FE type as the original.
To determine the FE type of the generated spans, we train an FE type classifier on FrameNet by finetuning SpanBERT, the state-of-the-art model for span classification \cite{Joshi2019SpanBERTIP}.\footnote{Our SpanBERT FE classifier attains 95\% accuracy on the standard FrameNet 1.7 splits; see \autoref{app:fe-classifier} for details.}
We use a strict filtering criterion: remove all generations where the FE classifier detects even a single FE type inconsistency, i.e. only retain instances with perfect FE fidelity.

\subsection{Intrinsic Evaluation of Generations} 
\label{sec:intrinsic}

We evaluate our generated frame-semantic annotations against those from \citet{Pancholy2021SisterHD}, before and after filtering (\S\ref{sec:method-filtering}).
We consider three metrics: perplexity under Llama-2-7B \cite{Touvron2023Llama2O} for overall fluency, FE fidelity, and human acceptance.
We randomly sampled 1000 LUs without annotations under FrameNet and used our generation framework to generate one instance each for these LUs. 
For human acceptability, we perform fine-grained manual evaluation on 200 examples sampled from the generated instances.\footnote{Human evaluation is mainly conducted by the first author of this work.
These annotations were validated by two independent volunteers unfamiliar with generated data evaluating the same examples from \gptframefe, where the ratings differ by only 1\% from our primary ratings. 
This suggests a consistent rating quality across different observers. } 
We deem an example acceptable if the FE spans semantically align with the target LU and preserve the FE role definitions under FrameNet.
We provide a qualitative analysis of generated examples in \autoref{app:human-eval}.

Results in \autoref{tab:intrinsic} shows that our filtering approach---designed for perfect FE fidelity---improves performance under the other two metrics.
Compared to rule-based generations from \citet{Pancholy2021SisterHD}, our filtered generations fare better under both perplexity and human acceptability, indicating improved fluency and semantic consistency.
Most importantly, models incorporating semantic information, i.e., {FE-conditioned} and {Frame+FE-conditioned} models, achieve higher human acceptance and generally lower perplexity compared to their no-conditioning counterparts, signifying that semantic cues improve both fluency and semantic consistency.
Even before filtering, FE fidelity increases with the amount of semantic conditioning, indicating the benefits of structure-based conditioning.
We also provide reference-based evaluation in \autoref{app:intrinsic-sim}.

\section{Augmenting Data for Frame-SRL}
\label{sec:augmentation}

Beyond improving FrameNet coverage, we investigate the extrinsic utility of our generations as training data to improve the frame-SRL task, which involves identifying and classifying FE spans in sentences for a given frame-LU pair.
Here, we consider a modified Frame-SRL task, which considers gold-standard frames and LUs, following \citet{Pancholy2021SisterHD}.
This remains a challenging task even for powerful models like GPT-4, which achieves a test F1 score of only 0.228 in contrast to \citet{Lin2021AGN}'s state-of-the-art F1 score of 0.722.
For experimental ease, we fine-tune a SpanBERT model on FrameNet's full-text data as our parser\footnote{This parser obtains an F1 score of 0.677, see \autoref{tab:aug-results}.} and avoid using existing parsers due to their reliance on weaker, non-Transformer architectures \cite{Swayamdipta2017FrameSemanticPW}, complex problem formulation \cite{Lin2021AGN}, or need for extra frame and FE information \cite{Zheng2022QueryYM}.

As a pilot study, we prioritize augmenting the training data with verb LUs with F1 scores below 0.75 on average.  
This serves as an oracle augmenter targeting the lowest-performing LUs in the test set. 
For the generation of augmented data, we use our top-performing models within \tfive and \gpt models according to human evaluation: \tfivefe and \gptframefe models.
Of 2,295 LUs present in the test data, 370 were selected for augmentation, resulting in 5,631 generated instances. 
After filtering, we retain 4,596 instances from \gptframefe and 4,638 instances from \tfivefe. 
Additional experiments using different augmentation strategies on subsets of FrameNet are in \autoref{app:aug-results-verb}.

\begin{table}[h!]
\centering
\resizebox{0.499\textwidth}{!}{
\begin{tabular}{lcc}
\toprule
& \bf All LUs F1 & \bf Aug. LUs F1 \\ \midrule
Unaugmented & 0.677 $\pm$ 0.004  & 0.681 $\pm$ 0.012 \\ 
Aug. w/ \texttt{T5} | FE & 0.683 $\pm$ 0.000   & 0.682 $\pm$ 0.006 \\
Aug. w/ \texttt{GPT-4} | Frame+FE & 0.684 $\pm$ 0.002 & 0.677 $\pm$ 0.010 \\ 
\bottomrule
\end{tabular}
}
\caption{F1 score of all LUs and augmented LUs under unaugmented setting, augmented settings with generations from \tfivefe and \gptframefe, averaged across 3 random seeds.}
\label{tab:aug-results}
\end{table}
\autoref{tab:aug-results} shows the Frame-SRL performance, with and without data augmentation on all LUs and on only the augmented LUs.
Despite the successes with human acceptance and perplexity, our generations exhibit marginal improvement on overall performance, and even hurt the performance on the augmented LUs.
We hypothesize that this stagnation in performance stems from two factors: (1) the phenomenon of diminishing returns experienced by our Frame-SRL parser, and (2) the limited diversity in augmented data.
Apart from the newly generated FE spans, the generated sentences closely resemble the original, thereby unable to introduce novel signals for frame-SRL; see \autoref{app:mult-candidates} and \autoref{app:gen-diversity} for more experiments on generation diversity. 
We speculate that \citet{Pancholy2021SisterHD}'s success with data augmentation despite using only sister LU replacement might be attributed to use of a weaker parser \cite{Swayamdipta2017FrameSemanticPW}, which left more room for improvement.

\subsection{Augmenting Under Low-Resource Setting}
\label{app:augment-diminishing}

To further investigate our failure to improve frame-SRL performance via data augmentation, we simulate a low-resource scenario and conduct experiments using increasing proportions of FrameNet training data under three settings: (1) training our SRL parser with full-text data, (2) training our SRL parser with both full-text and lexicographic data (which contains 10x more instances), and (3) training an existing frame semantic parser \cite{Lin2021AGN}\footnote{\citet{Lin2021AGN} break frame-SRL into three subsequent sub-tasks:  target identification, frame identification, and SRL, contributing to worse overall performance.} 
with full-text data, to control for the use of our specific parser.

\begin{figure}[!h]
\centering
\includegraphics[width=0.499\textwidth]{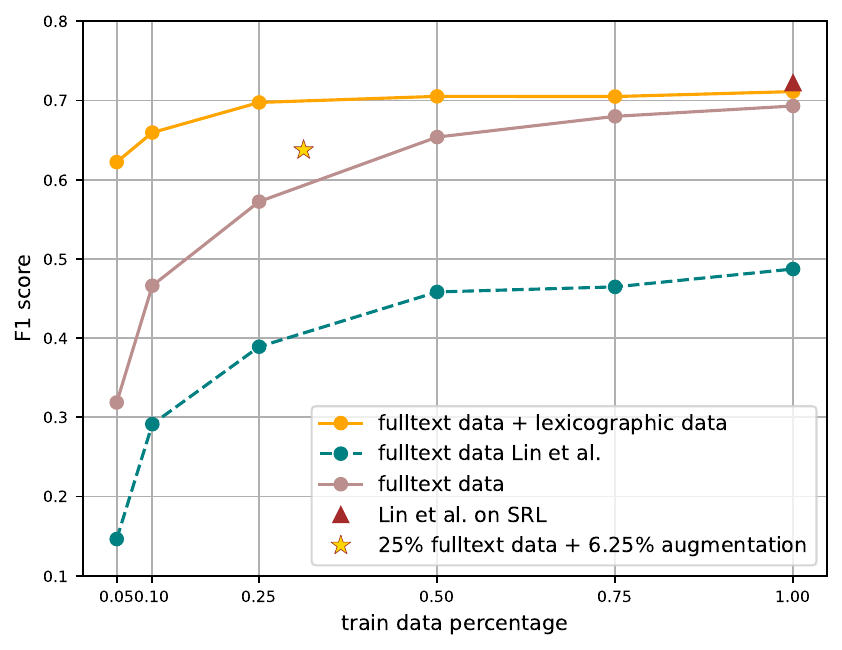}
\caption{Learning curves for our frame-SRL model and \citet{Lin2021AGN}'s end-to-end parser show diminishing returns on adding more human-annotated training data. The triangle marker denotes the performance of \citet{Lin2021AGN}'s parser on SRL with gold frame and LU.
}
\label{figure:scalability}
\end{figure}

\autoref{figure:scalability} shows that parsers across all three settings exhibit diminishing returns, especially on the second setting, which utilizes the largest training set. 
This indicates that there seems to be little room for improvement in frame-SRL, even with human annotated data.

Following our learning curves, we further evaluate the utility of our generations without the influence of diminishing returns, by performing data augmentation in a low-resource setting.
Specifically, we augment 25\% of the full-text training data with an additional 6.25\% of data generated using our method. 
As demonstrated in \autoref{figure:scalability}, the performance of the model in this scenario not only exceeds that of the 25\% dataset without augmentation but the results of the 25\% dataset augmented with 6.25\% of human-annotated data. 
This showcases the high utility of our generations for targeted data augmentation in a low-resource setting.

\section{Related Work}
\label{sec:related}

\paragraph{Data Augmentation for FrameNet}
While FrameNet annotations are expert annotated for the highest quality, this also limits their scalability. 
In an effort to improve FrameNet's LU coverage, \citet{pavlick-etal-2015-framenet} proposes increasing the LU vocabulary via automatic paraphrasing and crowdworker verification, without expanding the  lexicographic annotations.
Others address this limitation by generating annotations through lexical substitution \cite{Anwar2023TextAF} and predicate replacement \cite{Pancholy2021SisterHD}; neither leverages the generative capabilities of LLMs, however.

\paragraph{Controlled Generation}
Other works have explored using semantic controls for generation tasks.
\citet{ou-etal-2021-infillmore} propose FrameNet-structured constraints to generate sentences to help with a story completion task. \citet{Ross2021TailorGA} studied controlled generation given target semantic attributes defined within PropBank, somewhat coarse-grained compared to FrameNet. 
Similarly, \citet{Ye2024LLMDADA} employ the rewriting capabilities of LLMs to generate semantically coherent sentences that preserve named entities for the Named Entity Recognition task. 
\citet{Guo2022GENIUSSL} introduced GENIUS, a novel sketch-based language model pre-training approach aimed at reconstructing text based on keywords or sketches, though not semantic structures; this limits its effectiveness in capturing the full context.

\section{Conclusion} 
Our study provides insights into the successes and failures of LLMs in manipulating FrameNet's linguistic structures. 
When conditioned on semantic information, LLMs show improved capability in producing semantically annotated sentences, indicating the value of linguistic structure in language generation. 
Under a low-resource setting, our generated annotations prove effective for augmenting training data for frame-SRL. Nevertheless, this success does not translate to a high-resource setting, echoing challenges reported in applying LLMs to other flavors of semantics \cite{Bai2023ConstituencyPU,Lin2023ChatGPTIA,Ettinger2023YouAA}. 
These outcomes underline the need for further exploration into how LLMs can be more effectively employed in automating linguistic structure annotation. 

\section*{Acknowledgements}
We thank the anonymous reviewers and area chairs for valuable feedback.
This work benefited from several fruitful discussions with Nathan Schneider, Miriam R. L. Petruck, Jena Hwang, and many folks from the USC-NLP group.
We thank Ziyu He for providing additional human evaluation on generated annotations.
This research was partly supported by the Allen Institute for AI and an Intel Rising Stars Award.

\section*{Limitations}
While our work contributes valuable insights into LLMs' capabilities towards semantic structure-conditioned generation, we acknowledge certain limitations. 
First, our research is exclusively centered on the English language. 
This focus restricts the generalizability of our findings to other languages, which likely present unique linguistic structures with associated semantic complexity. 
The exploration of LLMs' capabilities in linguistic structures manipulation and generation in languages other than English remains an open direction for future research.

Moreover, we do not consider the full complexity of the frame semantic role labeling task, which also considers target and frame identification. Even for the argument identification task, we use an oracle augmentation strategy. 
Despite this relaxed assumption, the generations had limited improvement in performance, except in low-resource settings, where targeted data augmentation proved more effective. This indicates potential for improvement in scenarios with limited annotated data but highlights the need for further research in diverse and complex settings.

\section*{Ethics Statement}
We recognize the inherent ethical considerations associated with utilizing and generating data via language models. 
A primary concern is the potential presence of sensitive, private, or offensive content within the FrameNet corpus and our generated data. 
In light of these concerns, we carefully scrutinize the generated sentences during the manual analysis of the 200 generated examples and do not find such harmful content. 
Moving forward, we are committed to ensuring ethical handling of data used in our research and promoting responsible use of dataset and language models.

\clearpage
\appendix
\section{FrameNet Statistics}
\label{app:fn-pos-stats}

\subsection{Distribution of Lexical Units}
\autoref{tab:pos-stats} illustrates a breakdown of FrameNet corpus categorized by the POS tags of the LUs. 
Specifically, we report the number of instances and the average count of candidate FEs per sentence, corresponding to LUs of each POS category. 
The two predominant categories are verb (v) LUs and noun (n) LUs, with verb LUs exhibiting a higher average of candidate FE spans per sentence compared to noun LUs.
\begin{table}[h!]
\centering
\small
\begin{tabular}{lrrrr}
\toprule
\bf LU POS  & \bf \# Inst. & \bf \# FEs  & \bf \# C. FEs  & \bf \# Cd. FEs \\ \hline
v  & 82710  & 2.406  & 1.945  & 1.354  \\
n  & 77869  & 1.171  & 0.675  & 0.564  \\
a  & 33904  & 1.467  & 1.211  & 1.025  \\
prep  & 2996 & 2.212  & 2.013  & 1.946  \\
adv  & 2070  & 1.851  & 1.717  & 1.655  \\
scon  & 758  & 1.906  & 1.883  & 1.883  \\
num  & 350  & 1.086  & 0.929  & 0.549  \\
art  & 267  & 1.547  & 1.543  & 1.408  \\
idio  & 105  & 2.162  & 1.933  & 1.486  \\
c  & 69  & 1.957  & 0.841  & 0.826  \\
\bottomrule
\end{tabular}
\caption{Number of instances and average number of all, core, and candidate FE spans per sentence, categorized by POS tags of LUs in FrameNet. \textbf{C. FEs} represents Core FEs and \textbf{Cd. FEs} represents Candidate FEs.}
\label{tab:pos-stats}
\end{table}

\subsection{Replacement of non-verb LUs}
\autoref{tab:noun-replace} shows several examples of non-verb LU replacement, where the resulting sentences mostly preserve semantic consistency.
Given the extensive number of annotated verb LUs available for LU replacement and candidate FEs per sentence for masking and subsequent structure-conditioned generation, our generation methodology is primarily applied to verb LUs.
\begin{table}[]
\centering
\small
\begin{tabular}{p{16mm}p{20mm}p{30mm}}
\toprule
\textbf{Frame}     & \textbf{LU}     & \textbf{Sentence} \\ \hline
Leadership         & king.n (rector.n)   & No prior Scottish \textcolor{teal}{king} \textcolor{orange}{(rector)} claimed his minority ended at this age.  \\[0.05in]
Sounds             & tinkle.n (yap.n)    & Racing down the corridor, he heard the \textcolor{teal}{tinkle} \textcolor{orange}{(yap)} of metal hitting the floor.  \\[0.05in]
Body\_part         & claw.n (back.n)     & A cat scratched its \textcolor{teal}{claws} \textcolor{orange}{(back)} against the tree.  \\[0.05in]
Disgraceful \_situation   & shameful.a (disgraceful.a)   & This party announced his \textcolor{teal}{shameful} \textcolor{orange}{(disgraceful)} embarrassments to the whole world .    \\[0.05in]
Frequency          & always.adv (rarely.adv)   & The temple is \textcolor{teal}{always} \textcolor{orange}{(rarely)} crowded with worshippers .   \\[0.05in]
Concessive         & despite.prep (in spite of.prep)   & \textcolor{teal}{Despite} \textcolor{orange}{(In spite of)} his ambition , Gass ' success was short-lived .   \\[0.05in]
Conditional \_Occurrence   & supposing.scon (what if.scon)   & So , \textcolor{teal}{supposing} \textcolor{orange}{(what if)} we did get a search warrant , what would we find ?   \\
\bottomrule
\end{tabular}
\caption{Example sentences of non-verb LUs where semantic consistency is preserved after sister LU replacement. The original LU is in teal and the replacement LU is in orange and parentheses.}
\label{tab:noun-replace}
\end{table}

\subsection{Full-Text and Lexicographic Data}
\label{app:fn-splits}
\autoref{tab:train-test-split} shows the distribution of the training, development, and test datasets following standard splits on FrameNet 1.7 from prior work \cite{Kshirsagar2015FrameSemanticRL, Swayamdipta2017FrameSemanticPW, Peng2018LearningJS, Zheng2022QueryYM}.
Both the development and test datasets consist exclusively of full-text data, whereas any lexicographic data, when utilized, is solely included within the training dataset.
\begin{table}[h]
\centering
\begin{tabular}{lr}
\toprule
\bf Dataset Split  & \bf Size    \\ \midrule
Train (full-text + lex.)  & 192,364  \\
Train (full-text)  & 19,437   \\
Development  & 2,272 \\
Test   & 6,462  \\ 
\bottomrule
\end{tabular}
\caption{Training set size with and without lexicographic data, development set size, and test set size in FrameNet 1.7. 
}
\label{tab:train-test-split}
\end{table}
Since our generation approach is designed to produce lexicographic instances annotated for a single LU, when augmenting fulltext data (\S\ref{sec:augmentation}), we break down each fulltext example by annotated LUs and process them individually as multiple lexicographic examples.

\section{Details on Candidate FEs Selection}
\label{app:candidate-fe}
There are three criteria for determining a candidate FE span, i.e., FE Type Criterion, Ancestor Criterion, and Phrase Type Criterion. 
In preliminary experiments, we have conducted manual analysis on the compatibility of FE spans with replacement LUs on 50 example generations.
As demonstrated through the sentence in \autoref{figure:pipeline}, the FE Type criterion can effectively eliminate non-core FE that do not need to be masked, i.e., "Growing up" of FE type \fe{Time}. Also, the Phrase Type Criterion can identify the candidate FE "for breaking the rules", which is a prepositional phrase.  
Moreover, we find that FEs of Agent or Self-mover type describes a human subject, which is typically independent of the LU evoked in the sentence. 
Since FE types within the same hierarchy tree share similar properties, we exclude FEs of Agent and Self-mover types, as well as any FEs having ancestors of these types, from our masking process, as illustrated in \autoref{tab:cand-fe}. 
\begin{table}[]
\centering
\small
\begin{tabular}{p{40mm}p{25mm}}
\toprule
\textbf{Sentence After Replacement}  & \textbf{FE Type}     \\ \hline
\textcolor{teal}{She} was bending over a basket of freshly picked flowers , \textcolor{orange}{organizing} them to her satisfaction . & Agent (Agent)  \\ [0.35in]
\textcolor{teal}{The woman} got to her feet , \textcolor{orange}{marched} indoors , was again hurled out .  & Self\_mover (Self\_mover)   \\ [0.35in]
While \textcolor{teal}{some} \textcolor{orange}{presumed} her husband was dead , Sunnie refused to give up hope .  & Cognizer (Agent)  \\

\bottomrule
\end{tabular}
\caption{Example sentences after LU replacement with FEs of type \fe{Agent}, \fe{Self\_mover}, or their descendants, which are compatible with the new replacement LU. The ancestors of FE types are reported in parentheses. The FEs are shown in teal and the replacement LUs are shown in orange.}
\label{tab:cand-fe}
\end{table}

\section{Details on Span Generation}
\label{app:t5-tempplates}

\subsection{T5-large Fine-Tuning}
During the fine-tuning process of T5-large, we incorporate semantic information using special tokens, which is demonstrated in \autoref{tab:t5-format} through the example sentence in \autoref{figure:pipeline}.
T5 models are fine-tuned on full-text data and lexicographic data in FrameNet for 5 epochs with a learning rate of 1e-4 and an AdamW \cite{Loshchilov2017DecoupledWD} optimizer of weight decay 0.01. The training process takes around 3 hours on 4 NVIDIA RTX A6000 GPUs.

\subsection{GPT-4 Few-shot Prompting}
\label{app:gpt-prompts}
When instructing GPT-4 models to generate FE spans, we provide the task title, definition, specific instructions, and examples of input/output pairs along with explanations for each output, as demonstrated in \autoref{tab:prompt-example}.
\begin{table}[h!]
\centering
\small
\begin{tabular}{lp{40mm}}
\toprule
\textbf{Model}           & \textbf{Input} \\ \hline
No Conditioning   &  \texttt{Growing up, \color{brown}{<mask>}} are rewarded \texttt{\color{brown}{<mask>}}.             \\
FE-Conditioning & \texttt{ Growing up, \color{teal}{<FE: Evaluee>} \texttt{\color{brown}{<mask>}} </FE: Evaluee>} are rewarded \texttt{\color{teal}{<FE: Reason>} \texttt{\color{brown}{<mask>}} </FE: Reason>}.            \\ 
Frame-FE-Conditioning & \texttt{Growing up, \color{orange}{<Frame: Rewards\_and\_Punishments + \color{teal}{FE: Evaluee}\color{orange}>}} \texttt{\color{brown}{<mask>}} \texttt{\color{orange}{</Frame: Rewards\_and\_Punishments + \color{teal}{FE: Evaluee}\color{orange}>}} are rewarded \texttt{\color{orange}{<Frame: Rewards\_and\_Punishments + \color{teal}{FE: Reason}\color{orange}>}} \texttt{\color{brown}{<mask>}} \texttt{\color{orange}{</Frame: Rewards\_and\_Punishments + \color{teal}{FE: Reason}\color{orange}>}}.            \\ 
\bottomrule
\end{tabular}
\caption{Template of finetuning T5 models on an example sentence.}
\label{tab:t5-format}
\end{table}
\begin{table*}[h]
\centering
\small
\begin{tabular}{c|p{125mm}}
\toprule
Title  & \texttt{Sentence completion using frame elements} \\ [0.05in]
Definition  & \texttt{You need to complete the given sentence containing one or multiple blanks (<mask>). \textcolor{teal}{Your answer must be of the frame element type specified in FE Type.}} \\ [0.05in]
Example Input  & \texttt{\textcolor{orange}{\textbf{Frame}: Rewards\_and\_Punishments.} \textbf{Lexical Unit}: discipline.v. \textbf{Sentence}: Growing up, <mask> are disciplined <mask>. \textcolor{teal}{\textbf{FE Type}: Evaluee, Reason.}} \\ [0.05in]
Example Output  & \texttt{boys, for breaking the rules} \\ [0.05in]
Reason  & \texttt{\textcolor{orange}{The frame "Rewards\_and\_Punishments" is associated with frame elements "Evaluee" and "Reason".} The answer "boys" fills up the first blank \textcolor{teal}{because it is a frame element (FE) of type "Evaluee"}. The answer "for breaking the rules" fills up the second blank \textcolor{teal}{because it is an FE of type "Reason"}.} \\ [0.05in]
Prompt  & \texttt{Fill in the blanks in the sentence based on the provided \textcolor{orange}{frame,} lexical unit \textcolor{teal}{and FE type}. Generate the spans that fill up the blanks ONLY. Do NOT generate the whole sentence or existing parts of the sentence. Separate the generated spans of different blanks by a comma. Generate the output of the task instance ONLY. Do NOT include existing words or phrases before or after the blank.} \\ [0.05in]
Task Input  & \texttt{\textcolor{orange}{\textbf{Frame:} Experiencer\_obj.} \textbf{Lexical Unit}: please.v. \textbf{Sentence}: This way <mask> are never pleased <mask> . \textcolor{teal}{\textbf{FE Type}: Experiencer, Stimulus.}} \\ [0.05in]
Task Output & \\
\bottomrule
\end{tabular}
\caption{Example prompts for GPT-4 models. Texts in green only appear in \texttt{FE-Conditioning} and \texttt{Frame-FE-Conditioning} models. Texts in orange only appear in \texttt{Frame-FE-Conditioning} models. }
\label{tab:prompt-example}
\end{table*}

\section{FE Classifier Training Details}
\label{app:fe-classifier}
Our classifier operates on the principle of classifying one FE span at a time. 
In cases where multiple FE spans are present within a single sentence, we split these into distinct instances for individual processing. 
For each instance, we introduce special tokens—\texttt{<LU\_START>} and \texttt{<LU\_END>}—around the LU, and \texttt{<FE\_START>} and \texttt{<FE\_END>} around the FE span. 
Additionally, the name of the evoked frame is appended to the end of the sentence. 
To train our classifier to effectively discern valid FE spans from invalid ones, we augment training data with instances where randomly selected word spans are labeled as “Not an FE”, constituting approximately 10\% of the training data. 
The FE classifier is fine-tuned on full-text data and lexicographic data for 20 epochs with a learning rate of 2e-5 and an AdamW optimizer with weight decay 0.01. 
The training process takes around 4 hours on 4 NVIDIA RTX A6000 GPUs. 

\section{Human evaluation of generated examples}
\label{app:human-eval}
We perform fine-grained manual analysis on 200 generated sentences to evaluate the quality of model generations based on two criteria: (1) sentence-level semantic coherence and (2) preservation of original FE types. We present 10 example sentences from the overall 200 in \autoref{tab:human-eval}.
\begin{table*}[]
\centering
\small
\begin{tabular}{p{18mm}p{18mm}p{40mm}p{20mm}p{20mm}p{10mm}}
\toprule
\bf Frame  & \bf LU  & \bf Sentence  & \bf Original FEs  & \bf \gptfe  & \bf Human Eval.  \\ \hline
Verification  & verify.v (confirm.v)  & The bank, upon \textcolor{orange}{confirming} \textcolor{teal}{\fe{<Unconfirmed\_content>}}, released the goods to the customer.   & compliance with the terms of the credit   & the transaction details  & $\checkmark$ $\checkmark$  \\[0.5in]

Distributed \_position  & blanket.v (line.v)  & \textcolor{teal}{\fe{<Theme>}} \textcolor{orange}{lines} \textcolor{teal}{\fe{<Location>}} and the lake is covered with ice.   & snow many feet deep, the land   & the first snowfall, the shore   & $\checkmark$ $\checkmark$  \\[0.3in]

Being\_located   & sit.v (stand.v)   & Against the left-hand wall nearest to the camera are three storage shelves; \textcolor{teal}{\fe{<Theme>}} \textcolor{orange}{stands} \textcolor{teal}{\fe{<Location>}}.   & a lidless unvarnished coffin in the process of construction, on the middle shelf   & a tall vase, on the top shelf   & $\checkmark$ $\checkmark$   \\[0.65in]

Evoking   & conjure.v (evoke.v)   & A name like Pauline Gascoyne inevitably \textcolor{orange}{evoke} \textcolor{teal}{\fe{<Phenomenon>}}.   & an image of a bimbo Gazza in a GTi   & memories of a bygone era   & $\checkmark$ $\checkmark$   \\[0.4in]

Event   & happen.v (take place.v)   & Jamaicans appear to worry little about the future; sometimes it seems that they worry little even about what \textcolor{orange}{takes place} \textcolor{teal}{\fe{<Time>}}.   & in the next few minutes   & tomorrow   & $\checkmark$ $\checkmark$   \\[0.5in]

Self\_motion   & climb.v (walk.v)   & My mother parked her bicycle in the shoulder and took my hand, and we \textcolor{orange}{walked} \textcolor{teal}{\fe{<Goal>}}.   & to the top of the hill   & to the park   & $\checkmark$ $\checkmark$   \\[0.4in]

Process\_materials   & stain.v (process.v)   & If you accidentally \textcolor{orange}{process} \textcolor{teal}{\fe{<Material>}} \textcolor{teal}{\fe{<Alterant>}}, leave it for a week or two.   & walls, with woodworm fluid   & the wood, too much   & $\checkmark$ $\times$   \\[0.4in]

Self\_motion   & creep.v (make.v)   & Matilda took the knife she had been eating with, and all four of them \textcolor{orange}{make} \textcolor{teal}{\fe{<Path>}}.   & towards the dining-room door   & their way to the living room   & $\checkmark$ $\times$   \\[0.4in]

Hunting   & hunt.v (fish.v)    & \textcolor{teal}{\fe{<Food>}} too were mercilessly \textcolor{orange}{fished} and often left, plucked and dying, where the sealers found them.   & The albatrosses   & The penguins   & $\times$ $\checkmark$   \\[0.5in]

Change\_position \_on\_a\_scale   & dip.v (rise.v)   & \textcolor{teal}{\fe{<Attribute>}} \textcolor{orange}{rose} \textcolor{teal}{\fe{<Final \_value>}} in the summer, but has recently climbed above \$400 and last night was nudging \$410.   & The price per ounce, below \$360   & The price, to \$410   & $\times$ $\checkmark$   \\[0.5in]

\bottomrule
\end{tabular}
\caption{Example Generations of \gptfe, our best model according to human acceptance. The two marks in human evaluation represent whether the generations satisfy the two criteria individually: (1) sentence-level semantic coherence and (2) preservation of all FE types. A sentence is deemed acceptable only when it satisfies both criteria. The new replacement LUs are presented in orange or parentheses. Masked FE spans are presented in teal and their corresponding FE types in angle brackets.}
\label{tab:human-eval}
\end{table*}

\section{Intrinsic Evaluation on FrameNet Test Data}
\label{app:intrinsic-sim}
To evaluate the quality of generated sentences on reference-based metrics such as ROUGE \cite{Lin2004ROUGEAP} and BARTScore \cite{Yuan2021BARTScoreEG}, we perform \S\ref{sec:method-masking} and \S\ref{sec:method-generation} on the test split of FrameNet 1.7 with verb LUs. 
As observed in \autoref{tab:intrinsic-sim}, the \tfivefe model surpasses others in ROUGE scores, signifying superior word-level precision, while \gpt achieves the highest BARTScore, indicating its generated sentences most closely match the gold-standard FE spans in terms of meaning. 
For reference-free metrics, \gptfe performs well in both log perplexity and FE fidelity, showcasing its ability to produce the most fluent and semantically coherent generations.

\begin{table*}[h!]
\centering
\begin{tabular}{cccccc}
\toprule
& BARTScore & ROUGE-1 & ROUGE-L & Perp. & FE Fid.               \\ \hline
Human  & -  & -  & -  & 4.82 & -   \\ 
\hline
\texttt{T5} & -5.939 & 0.301 & 0.298 & 447.874 & 0.829    \\
\texttt{T5} | FE & -5.922 & \textbf{0.318} & \textbf{0.316} & 434.231 & 0.840    \\
\texttt{T5} | Frame + FE  & -6.179 & 0.276 & 0.274 & 441.639 & 0.843    \\ \hline
\texttt{GPT-4}  & \textbf{-4.060} & 0.228 & 0.227 & 85.820 & 0.880    \\
\texttt{GPT-4} | FE  & -4.336 & 0.218 & 0.217 & \textbf{82.977} & \textbf{0.930}    \\
\texttt{GPT-4} | Frame + FE  & -4.395 & 0.210 & 0.209 & 87.548 & 0.929    \\
\bottomrule
\end{tabular}
\caption{Log BARTScore, ROUGE scores and perplexity of generations on FrameNet test set without LU replacement.}
\label{tab:intrinsic-sim}
\end{table*}

\section{More on Augmentation Experiments}
\label{app:aug-results-verb}
\subsection{Experiments using Non-oracle Augmentation Strategy}
To evaluate the robustness and generalizability of our model under realistic conditions, we employed an augmentation strategy similar to that used by \citet{Pancholy2021SisterHD}. Specifically, we remove all annotated sentences of 150 randomly selected verb LUs from the full text training data and train our baseline parser using the remaining training data. Our full model was trained on instances of the 150 verb LUs re-generated by our framework along with the data used to train the baseline model. As a result, the test F1 scores for the baseline model and full model were 0.689 and 0.690, respectively, which echos the lack of significant improvement using the oracle augmentation strategy.

\subsection{Experiments on Verb-only Subset}
Since our generation method mainly focuses on augmenting verb LUs, we conduct additional augmentation experiments using a subset of FrameNet that includes only verb LU instances.  
To ensure model performance on a subset of data, we incorporate lexicographic data with verb LUs into our training set, resulting in a training set enriched with 80.2k examples, a development set comprising approximately 600 examples, and a test set containing about 2k examples. 
We experimented with different augmentation percentages both with and without filtering, as shown in \autoref{tab:aug-results-verb}. 
We use an oracle augmenter to augment LUs inversely proportional to their F1 scores from the unaugmented experiments. 
To expand coverage on more LUs during augmentation, we augment all LUs rather than limiting to those with F1 scores below 0.75.
Although the improvements are marginal, the outcome from filtered augmentations is generally better than those from their unfiltered counterparts.
\begin{table}[h!]
\centering
\resizebox{0.495\textwidth}{!}{
\begin{tabular}{lcc}
\toprule
& \bf All LUs F1 & \bf Aug. LUs F1 \\ \hline
Unaugmented & 0.751  & 0.779  \\ 
5\% Aug. w/o filter & 0.745   & 0.778  \\
5\% Aug. w/ filter & 0.752  &\textbf{0.781}  \\ 
25\% Aug. w/o filter & 0.752   & 0.776 \\
25\% Aug. w/ filter & \textbf{0.753}  & \textbf{0.781}  \\ 
\bottomrule
\end{tabular}
}
\caption{F1 score of all verb LUs and augmented LUs in  augmentation experiments using different percentages of augmentations generated by \texttt{T5} | FE with and without filtering, compared to baseline results without data augmentation. Best results are in boldface}
\label{tab:aug-results-verb}
\end{table}

\subsection{Experiments on Multiple Candidate Generations}
\label{app:mult-candidates}
In the main experiments conducted in this paper, we generated one instance for each LU-sentence pair. However, instances could be filtered out due to inconsistent FE spans, which could hurt generation diversity. To address this, we further experimented with generating three candidate instances for each LU-sentence pair to improve generation coverage.

Specifically, we augmented the full-text training data by 25\% under both the 1-candidate and 3-candidate settings. However, as shown in \autoref{tab:mult-candidates}, generating three candidates did not lead to performance improvements in the F1 score. This suggests that simply increasing the number of generated candidates may not be sufficient to enhance generation diversity. Future work may need to explore more effective strategies to improve the diversity of generated data.

\begin{table}[h!]
\centering
\begin{tabular}{lc}
\toprule
& \bf All LUs F1  \\ \hline
Unaugmented & 0.693  \\ 
1-candidate & 0.688   \\
3-candidate & 0.673   \\ 
\bottomrule
\end{tabular}
\caption{F1 score of SRL parsers trained on unaugmented data and augmented data generated by \texttt{T5} | FE under 1-candidate and 3-candidate strategies.}
\label{tab:mult-candidates}
\end{table}

\section{Effect of Filtering on Generation Diversity}
\label{app:gen-diversity}
\begin{table}[h!]
\centering
\resizebox{0.495\textwidth}{!}{
\begin{tabular}{lcc}
\toprule
& \bf Before Filtering & \bf After Filtering \\ \hline
Human & 0.298  & -  \\ 
\tfive & 0.302   & 0.278  \\
\tfivefe & 0.295  & 0.277  \\ 
\tfiveframefe & 0.295   & 0.271 \\
\gpt & 0.270   & 0.249  \\
\gptfe & 0.268  &0.246  \\ 
\gptframefe & 0.271   & 0.253 \\
\bottomrule
\end{tabular}
}
\caption{Self-BLEU scores of the 1000 instances created in \S\ref{sec:intrinsic} before and after filtering.}
\label{tab:gen-diversity}
\end{table}
To examine the effect of filtering on the diversity of generated data, we have conducted experiments to compute the Self-BLEU scores to measure diversity for the same 1,000 instances discussed in \S\ref{sec:intrinsic}. A lower Self-BLEU score indicates higher diversity, as it signifies less overlap within the generated texts. As demonstrated in \autoref{tab:gen-diversity}, the diversity of the generated candidates increases after applying the filter, even surpassing the diversity of the original instances created by humans. This substantiates the effectiveness of our filtering process in enhancing the variability and quality of the generated sentences.

\end{document}